\documentclass[letterpaper, 10 pt, conference]{ieeeconf}

\IEEEoverridecommandlockouts
\usepackage[noadjust]{cite}
\usepackage{longtable}
\usepackage{amsmath}
\usepackage{graphicx,epstopdf,amssymb}
\usepackage[dvips]{epsfig}
\usepackage{color}
\usepackage{bbm}
\usepackage{dsfont}
\usepackage{enumerate}
\PassOptionsToPackage{hyphens}{url}
\usepackage{hyperref}
\usepackage{algpseudocode}
\usepackage{algorithm}
\usepackage{fixltx2e}
\usepackage{subcaption}
\usepackage[export]{adjustbox}

\newtheorem{lemma}{Lemma}
\newtheorem{remark}{Remark}

\newcommand{\comment}[1]{} 

\newcommand{\carre}{\hfill $\blacksquare$}
\newcommand{\carrew}{\hfill $\square$}

\newcommand{\SO}{\ensuremath{\text{SO}(3)}}

\newcommand{\Rn}[1][3]{\ensuremath{\mathds{R}^{#1}}}
\newcommand{\N}[1]{{\bar N{\hspace{-0.3mm}{}^{#1}_i}}^{\hspace{-1mm}\top}}
\newcommand{\normN}[1]{\|N^{#1}_i\|}

\newcommand{\gmin}{\ensuremath{g_\text{min}}}
\newcommand{\gmax}{\ensuremath{g_\text{max}}}

\title{\LARGE A Branch-and-Bound Algorithm for Checkerboard Extraction in Camera-Laser Calibration
	\vspace{-3mm}
	}
\author{
Alireza Khosravian, Tat-Jun Chin, Ian Reid \\
\thanks{The authors are with School of Computer Science, The University of Adelaide (\{alireza.khosravian, tat-jun.chin, ian.reid\} @adelaide.edu.au).}
\thanks{This work was supported by the Australian Research Council through the ARC Linkage Project LP140100946, the Center of Excellence in Robotic Vision CE140100016, and the Laureate Fellowship FL130100102. The authors greatly acknowledge the help and support of Maptek Pty. Ltd.}
\vspace{-1.2cm}
}

\begin{document}

\maketitle

\begin{abstract}
We address the problem of camera-to-laser-scanner calibration using a checkerboard and multiple image-laser scan pairs.  Distinguishing which laser points measure the checkerboard and which lie on the background is essential to any such system. 
We formulate the checkerboard extraction as a combinatorial optimization problem with a clear cut objective function. We propose a branch-and-bound technique that deterministically and globally optimizes the objective. Unlike what is available in the literature, the proposed method is not heuristic and does not require assumptions such as constraints on the background or relying on discontinuity of the range measurements to partition the data into line segments. The proposed approach is generic and can be applied to both 3D or 2D laser scanners as well as the cases where multiple checkerboards are present. We demonstrate the effectiveness of the proposed approach by providing numerical simulations as well as experimental results.   
\end{abstract}


\IEEEpeerreviewmaketitle

\section{Introduction}
Many robotics applications rely on a camera and a laser scanner, which are rigidly attached together, as the main sensors for performing localization, navigation, and mapping the surrounding environment. Effective fusion of laser measurements with camera images requires knowledge of the extrinsic transformation between the two sensors. 

The process of calibrating either a 2D or 3D laser scanner to a camera by computing the rigid transformation between their reference frames is known as \textit{camera-laser calibration}.  Various methods have been developed for this \cite{zhang2004extrinsic,scaramuzza2007extrinsic,vasconcelos2012minimal,napier2013cross,castorena2016autocalibration,pandey2012automatic,unnikrishnan2005fast,kassir2010reliable,mirzaei20123d,herrera2012joint,geiger2012automatic,wasielewski1995calibration}, including free toolboxes (e.g. \cite{kassir2010reliable} for 2D lasers and \cite{unnikrishnan2005fast,geiger2012automatic} for 3D lasers).
Although a few checkerboard free calibration methods have been developed in the literature (see e.g. \cite{scaramuzza2007extrinsic,castorena2016autocalibration,napier2013cross,pandey2012automatic}), still a very popular approach is to place one or multiple known targets (e.g. planar checkerboards) in the scene such that they are observed in both the image and the laser scans (see e.g. \cite{zhang2004extrinsic,vasconcelos2012minimal,herrera2012joint,geiger2012automatic,unnikrishnan2005fast,kassir2010reliable,mirzaei20123d,pandey2010extrinsic,herrera2011accurate,mei2006calibration}). We take the later approach in this paper. In this approach, one or multiple sets of laser-image pairs are collected. The checkerboard corners associated with each image are identified and the checkerboard plane is obtained in the camera coordinate frame. Then, the laser scan points that hit the checkerboard plane in each image are extracted and used to solve a nonlinear optimization that estimates the extrinsic transformation between the camera and laser scanner.

An important part of the process is \textit{checkerboard extraction}, in which we must determine which laser scan points fall on the checkerboard and which lie on the background. Manual extraction of those inliers is a time consuming and inaccurate process, although some popular toolboxes still rely on this manual extraction (see e.g. \cite{unnikrishnan2005fast}). Automatic checkerboard extraction has been poorly studied in the literature and \cite{kassir2010reliable} seemed to be the only reference that investigates this problem in details. Other references, only briefly discuss the checkerboard extraction as a block in the pipeline of the camera-laser calibration process \cite{mirzaei20123d,vasconcelos2012minimal,geiger2012automatic}. All of these methods are heuristic in nature, imposing assumptions on the environment or the camera-laser setup to identify possible candidate laser points that might correspond to the checkerboard. Those candidates are then passed to a randomised hypothesis-and-test search (e.g. RANSAC \cite{fischler1981random}) to eliminate the outliers and identify the correspondence of the laser points and the checkerboards. The imposed assumptions depend on whether a 2D or 3D laser scanner is employed and are specific to particular situations. For instance, where 2D laser scanner is employed, \cite{mirzaei20123d,kassir2010reliable} and \cite[Section 4]{vasconcelos2012minimal} rely on range discontinuities to partition the range measurements into line segments that might correspond to a planar checkerboard. Also \cite{kassir2010reliable} additionally assumes that the background is stationary in most of the scans and relies on the frequency of occurrence of range measurements. Assumptions on the minimum and maximum distance of the checkerboard to the laser scanner during different scans, have also been used to enhance the checkerboard extraction process \cite{geiger2012automatic}.

The heuristic methods have demonstrated good performance in some applications  \cite{kassir2010reliable,geiger2012automatic,vasconcelos2012minimal,mirzaei20123d,fischler1981random}. 
However, imposing assumptions limits the application of these methods to particular scenarios. Assumptions such as stationary background limit the usage of the developed techniques to the controlled environments, preventing their application to open spaces where the background could inevitably change due to moving objects. Also, in some practical scenarios, stationary known targets are used for calibration and the camera-laser pair (usually mounted on a vehicle) is moved around and collects the data. In such situations, the background does inevitably change. Relying on range discontinuities is limited to the cases where the checkerboard is not placed on the walls and there are not many other objects present in the scene that might cause unwanted range discontinuities. Little attention has been given in literature to the design of \textit{systematic} methods for checkerboard extraction based on rigorous theoretical foundations that can perform robustly in all practical situations.




Branch-and-bound (BnB) is a systematic method for discrete and combinatorial optimization. In the context of geometric model fitting, BnB is a popular technique in situations where outliers are present or measurement correspondences are unknown (see e.g. \cite{hartley2009global,breuel2003implementation,parra2015fast,olsson2009branch} and the references therein). Building on a rigorous theoretical foundation, this method systematically searches the state space for inliers that optimize an objective function and is guaranteed to find the globally optimal state. A BnB method has recently been developed for the hand-eye calibration problem \cite{heller2012branch,seo2009branch} and demonstrated successful application. To the best of our knowledge, the BnB method has not been applied to the camera-laser calibration problem prior to the current contribution.

In this paper, we propose a BnB technique for checkerboard extraction in the camera-laser calibration problem. Following \cite{zhang2004extrinsic}, we assume that checkerboard normals are obtained in the camera coordinate frame by processing the checkerboard corners. We propose an objective function that represents the number of laser points that fall within a box with a small distance to the checkerboard (inliers). A key contribution of the paper is to propose a tight upper bound for the objective function that ensures the BnB method finds the globally optimal inlier laser points. Our method is not heuristic and relies only on the underlying geometry of the camera-laser calibration problem without imposing any assumption or constraints on the environment. Hence it is quite general and performs robustly in all practical situations. It is applicable directly to both 2D and 3D laser scanners, to the case where multiple checkerboards are present in the scene, as well as the case where some of the checkerboards either do not fall or only partly fall in the field of view of the laser scanner or the camera. We demonstrate successful application of our methodology by providing simulation studies and experimental results. 

The inlier points extracted by our BnB approach can be used in any nonlinear optimization method to explicitly compute the extrinsic camera to laser transformation. The BnB method also provides a rough estimate of the extrinsic transformation which can be used for initializing the combined nonlinear optimization close to the optimal transformation, preventing it from local optimums.

\section{Problem formulation} \label{sec:problem:formulation}
\comment{Input from the range sensor is an unordered set of 3D points $p^l_{ij}$ which keeps the approach as
	generic as possible, applicable to both 2D and 3D laser scanners.}
Consider a camera-laser setup attached rigidly together. Denote the camera-fixed coordinate frame and the laser-fixed coordinate frame by $\{c\}$ and $\{l\}$ respectively. For simplicity, we assume that only one checkerboard is present in the scene. Nevertheless, the method presented in this paper is directly applicable to the case where multiple checkerboards are present (see Remark \ref{rem:multiple:checkerboards}). 
The representation of a point $p$ in the camera frame and laser scanner frame are denoted by $p^c \in \Rn$ and $p^l \in \Rn$, respectively. One has
\begin{align}
\label{equ:cam2laser} p^l = \Phi p^c + \Delta,
\end{align}
where $\Phi \in \SO$  and $\Delta \in \Rn$, respectively, denote the rotation and translation from $\{c\}$ to $\{l\}$ and represent the extrinsic camera-laser calibration. 

Assume that a calibration plane (e.g. a checkerboard) is placed in the environment such that it falls within the field of view of both the camera and the laser scanner. The laser scanner measures the 3D coordinate of the points of the environment in the laser-fixed frame. Some of these points fall in the plane of the checkerboard while others correspond to the points in the environment. For the purpose of this paper, we do not differentiate between 2D or 3D laser scanners as the theory presented here is applicable to both without any modification.

We assume the dimensions of the checkerboard are known and use standard image-based techniques to extract the location of intersection points in the image and find the camera intrinsic parameters as well as the pose parameters that relate the checkerboard's reference frame to the camera's. This task is easily accomplished using off-the-shelf software packages such as \cite{bouguet2010camera}. Changing the relative pose of the checkerboard with respect to (wrt.) the camera-laser setup (by moving either the checkerboard or the camera-laser setup), we collect $n$ sets of laser scan points $P_{ij}$ each contain $m$ points ($i=1,2,\ldots,n$ and $j=1,2,\ldots,m$). Each set of these points corresponds to a scene in which a checkerboard with a known extrinsic transformation wrt. the camera frame is present. The main objective of this paper is to develop an algorithm that takes $n$ sets of the laser scan points and their corresponding camera images and separates the laser points that correspond to the checkerboard (inliers) from those that belong to the background (outliers). 

\section{Checkerboard extraction using BnB} \label{sec:BnB}
\begin{figure} 
	\centering
	\includegraphics[height=4.3cm]{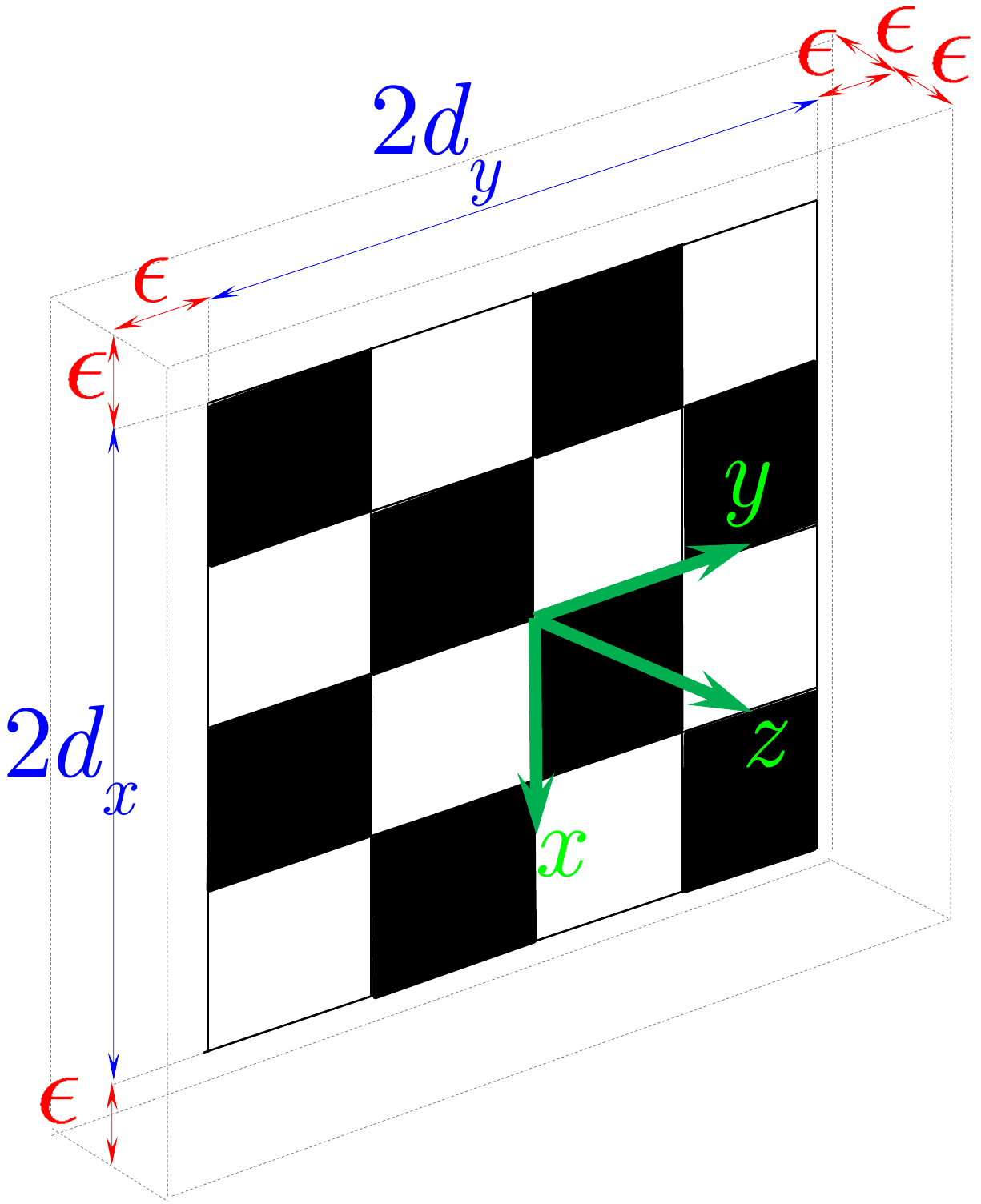}
	\vspace{-2mm}
	\caption{The checkerboard coordinate frame is shown in green. The checkerboard dimensions are $2d_x$ and $2d_y$. A laser point is considered an inlier if it falls inside the dashed box around the checkerboard whose dimensions are $(2d_x+2\epsilon) \times (2d_y+2\epsilon) \times 2\epsilon$.} \label{fig:checkerboard:box}
	\vspace{-5mm}
\end{figure}

We propose a branch-and-bound method (BnB) \cite{breuel2003implementation,olsson2009branch} for checkerboard extraction. Consider a coordinate frame attached to the center of the checkerboard with its z-axis normal to the checkerboard plane pointing outward (to the camera), its y-axis horizontal and its x-axis vertical, as depicted in Fig. \ref{fig:checkerboard:box}. Denoting the dimensions of the checkerboard along its $x$ and $y$ direction by $2 d_x$ and $2 d_y$, a laser point is considered an inlier if it falls inside a box around the calibration board whose dimensions along the $x$, $y$, and $z$ axes of the checkerboard are $2(d_x+\epsilon)$, $2(d_y+\epsilon)$, and $2\epsilon$, respectively, where $\epsilon>0$ is a user defined threshold
(see Fig. \ref{fig:checkerboard:box}). For the checkerboard in the $i$-th image, denoted by $N^x_i$, $N^y_i$, and $N^z_i$ vectors along the $x$, $y$, and $z$ axes of the checkerboard whose magnitude equals to the distance of the center of the camera frame to the $y-z$, $x-z$, and $x-y$ planes of the checkerboard coordinate frame, respectively. By detecting and processing the checkerboard corners, one can compute $N^x_i$, $N^y_i$, and $N^z_i$ in the camera coordinate frame
It is straight-forward to show that a laser point $p^l_{ij}$ falls inside the box of Fig. \ref{fig:checkerboard:box} iff the following conditions hold \cite{zhang2004extrinsic}.
\begin{subequations} \label{eq:box:condition}
\begin{align}
|\N{x} \Phi^\top (p^l_{ij}-\Delta)-\normN{x}| &< d_x + \epsilon,\\
|\N{y} \Phi^\top (p^l_{ij}-\Delta)-\normN{y}| &< d_y + \epsilon,\\
|\N{z} \Phi^\top (p^l_{ij}-\Delta)-\normN{z}| &< \epsilon,
\end{align}   
\end{subequations}
where $\bar N:=\frac{N}{\|N\|}$ denotes the unit vector parallel to $N$. Inspired by \cite{breuel2003implementation}, we propose the following objective function.
\begin{align}
\nonumber Q(\Phi,\Delta) \!=\!\! \sum_{i=1}^{n} \!\sum_{j=1}^{m} &\lfloor |\N{x} \Phi^\top (p^l_{ij}-\Delta)-\normN{x}| < d_x + \epsilon \rfloor\\
\nonumber & \times \!\lfloor |\N{y} \Phi^{\!\top} (p^l_{ij}\!-\Delta)\!-\!\normN{y}| < d_y + \epsilon \rfloor \\
\label{eq:Q} & \times\! \lfloor |\N{z} \Phi^{\!\top} (p^l_{ij}-\Delta)\!-\!\normN{z}| < \epsilon \rfloor,
\end{align}
where  $\lfloor .\rfloor$ is the indicator function. For a given transformation $(\Phi,\Delta)$, any point $p^l_{ij}$ that satisfies (\ref{eq:box:condition}) is an inlier and contributes $1$ to the summation in (\ref{eq:Q}). Hence, $Q$ represents the total number of the inliers for a given transformation. We wish to maximize $Q$ over the transformation $(\Phi,\Delta)$.

Algorithm \ref{alg:BnB} in the Appendix summarizes the BnB method used in this paper. We use the angle-axis representation to parametrize \SO. A 3D rotation is represented as a vector whose direction and norm specify the axis and angle of rotation, respectively \cite{hartley2009global,khosravian2010globally}. This simplifies the search space of $\Phi\in \SO$ to a ball with a radius of $\pi$. We enclose this ball with a box with a half length of $\pi$, denoted by $\mathbb{B}(\pi)$. In practice, usually an initial estimate of the bound on the amplitude of the rotation $\Phi$ is available. This initial estimate can be used to initialize the search space of rotations in the queue with a smaller box $\mathbb{B}(\delta_R)$ with $\delta_R \le \pi$. Similarly, the search space of translation is initialized with $\mathbb{B}(\delta_t)$ where $\delta_t$ represent an estimate of the maximum distance of the center of the camera from the center of the laser scanner.

The BnB method requires an upper bound for (\ref{eq:Q}). Our main contribution is to propose an upper bound in Section \ref{sec:upper:bound:function} (see Lemma \ref{lem:Qhat}). For branching, we simultaneously divide the boxes into eight sub-boxes, each with a half of the side length of the original box. Since we do the branching of the rotation and translation search spaces simultaneously at lines \ref{alg:line:branchR} and \ref{alg:line:brancht}  of Algorithm \ref{alg:BnB}, this branching yields the total of $8 \times 8=64$ new branches.
Note that the inliers that satisfy the condition (\ref{eq:box:condition}) are detected during evaluation of the objective function. Indexes of these inliers are returned as an output of the BnB method (see the line \ref{alg:line:inliers} of Algorithm \ref{alg:BnB}).
\section{Upper-bound function} \label{sec:upper:bound:function}
The key stage for employing the BnB method is to propose an upper-bound $\hat Q$ for (\ref{eq:Q}).
The following Lemma proposes an upper-bound and proves that it qualifies as a valid bounding function
for BnB. Detailed derivations are given in the proof.
\begin{lemma} \label{lem:Qhat}
	Consider the boxes $\mathbb{B}_R$ and $\mathbb{B}_t$ with the centers $\Phi_c \in \SO$ and $\Delta_c \in \Rn$ and the half side length of $\delta_R$ and $\delta_t$, respectively. Let
	\begin{align}
	 \nonumber \!\!\!\hat Q(\mathbb{B}_{\!R},\mathbb{B}_t\!) \!:=\!\! \sum_{i=1}^{n} \!\sum_{j=1}^{m} &\lfloor |\N{x} \Phi_c^{\!\top} (p^l_{ij}\!\!-\!\Delta_c)\!-\!\normN{x}| \!<\! d_x \!+\! \epsilon\!+\!\delta_{ij} \rfloor\\
	\nonumber & \!\!\!\!\!\!\!\!\!\times \!\!\lfloor |\N{y} \Phi_c^{\!\top} (p^l_{ij}\!-\!\Delta_c)\!-\!\normN{y}| \!<\! d_y \!+\! \epsilon\!+\!\delta_{ij} \rfloor \\
	 \label{eq:Qhat} & \!\!\!\!\!\!\!\!\!\times\!\! \lfloor |\N{z} \Phi_c^{\!\top} (p^l_{ij}\!-\!\Delta_c)\!-\!\normN{z}| \!<\! \epsilon\!+\!\delta_{ij} \rfloor
	\end{align}
where $\delta_{ij}$ is computed for each laser point $p^l_{ij}$ as
	\begin{align}
	\label{eq:deltaij} \delta_{ij}=\|p^l_{ij}-\Delta_c\|\sqrt{2(1-\cos(\sqrt{3}\delta_R))}+\sqrt{3} \delta_t.
	\end{align}
	The candidate upper-bound (\ref{eq:Qhat}) satisfies
	\begin{align}
	\label{eq:enq:Qhat} \hat Q(\mathbb{B}_R,\mathbb{B}_t) \ge \max_{\Phi \in \mathbb{B}_R,~\Delta \in \mathbb{B}_t} Q(\Phi,\Delta).
	\end{align} 
	Also, if $\mathbb{B}_R \times \mathbb{B}_t$ collapses to the single points $(\Phi,\Delta)$, we have $\hat Q(\mathbb{B}_R,\mathbb{B}_t) = Q(\Phi,\Delta)$.\carrew
\end{lemma} 

\textit{Proof of Lemma \ref{lem:Qhat}: }
Using the triangle inequality, we have
\begin{align}
\nonumber |&\bar N^\top \Phi_c^\top (p^l_{ij}-\Delta_c)-\|N\|| \le  |\bar N^\top \Phi^\top (p^l_{ij}-\Delta)-\|N\|| \\
\label{appendix:proof:lem:Qhat:tmp1} &+ |\bar N^\top \Phi_c^\top (p^l_{ij}-\Delta_c)-\bar N^\top \Phi^\top (p^l_{ij}-\Delta)|,
\end{align}
for any $N \!\in\! \Rn$. Again, applying triangle inequality yields
\begin{align}
\nonumber &|\bar N^\top \Phi_c^\top (p^l_{ij}-\Delta_c)-\bar N^\top \Phi^\top (p^l_{ij}-\Delta)| \\
\nonumber & \le |\bar N^\top \Phi^\top (\Delta-\Delta_c)| + |(\Phi_c \bar N - \Phi \bar N  )^\top (p^l_{ij}-\Delta_c) | \\
\label{appendix:proof:lem:Qhat:tmp2} &\le \|\Phi \bar N\| \|\Delta-\Delta_c\| + \|\Phi_c \bar N - \Phi \bar N \| \|p^l_{ij}-\Delta_c\| .
\end{align}
Since $\Delta \in \mathbb{B}_t$ and $\Delta_c$ is the center of the box $\mathbb{B}_t$ with the half side length $\delta_t$, we have $\|\Delta-\Delta_c\| \le \sqrt{3} \delta_t$. Also, since $\Phi \in \mathbb{B}_R$ and $\Phi_c$ is the center of the box $\mathbb{B}_R$ with the half side length $\delta_R$, and resorting to \cite[equation (6)]{parra2014fast} and noting that $\|\bar N\|=1$, we have $\|\Phi_c \bar N - \Phi \bar N \| \le \sqrt{2(1-\cos(\sqrt{3}\delta_R))}$. Combining these with (\ref{appendix:proof:lem:Qhat:tmp1}) and (\ref{appendix:proof:lem:Qhat:tmp2}) we have
\begin{align}
\nonumber |&\bar N^\top \Phi_c^\top (p^l_{ij}-\Delta_c)-\|N\|| \le  |\bar N^\top \Phi^\top (p^l_{ij}-\Delta)-\|N\|| \\
&+ \sqrt{3} \delta_t + \|p^l_{ij}-\Delta\|  \sqrt{2(1-\cos(\sqrt{3}\delta_R))}.
\end{align}
Hence, by (\ref{eq:deltaij}), $|\bar N^\top \Phi^\top (p^l_{ij}-\Delta)-\|N\|| \le \bar\epsilon$ implies $|\bar N^\top \Phi_c^\top (p^l_{ij}-\Delta_c)-\|N\|| -\delta_{ij} \le  \bar\epsilon$ for any $\bar{\epsilon} \ge 0$. Consequently,
$\lfloor|\bar N^\top \Phi^\top (p^l_{ij}-\Delta)-\|N\|| \le \bar\epsilon \rfloor \le \lfloor |\bar N^\top \Phi_c^\top (p^l_{ij}-\Delta_c)-\|N\|| \le  \bar\epsilon + \delta_{ij} \rfloor$.
The above derivations are valid if one replaces $N$ with $N^x_i$, $N^y_i$, or $N^z_i$ and substitutes $\bar{\epsilon}$ for $d_x+\epsilon$, $d_y+\epsilon$, or $\epsilon$, respectively. This proves (\ref{eq:enq:Qhat}). If $\mathbb{B}_R$ and $\mathbb{B}_t$ collapse to single points, we have $\Phi=\Phi_c$, $\Delta=\Delta_c$, and $\delta_R=\delta_t=0$ (thus $\delta_{ij}=0$). Substituting these into (\ref{eq:Qhat}) yields $\hat Q(\mathbb{B}_R, \mathbb{B}_t) = Q(\Phi,\Delta)$ and completes the proof. \carre

\subsection{Improving the tightness of the upper-bound}
Tightness of the upper bound function (\ref{eq:Qhat}) depends directly on the value of $\delta_{ij}$. The less conservative value of $\delta_{ij}$ is used, the tighter upper bound $\hat Q$ is obtained and BnB finds the global optimal value with less iterations. In this section, we tighten the upper bound function proposed in Section \ref{sec:upper:bound:function}.

\begin{figure} 
	\centering
	\includegraphics[height=5cm]{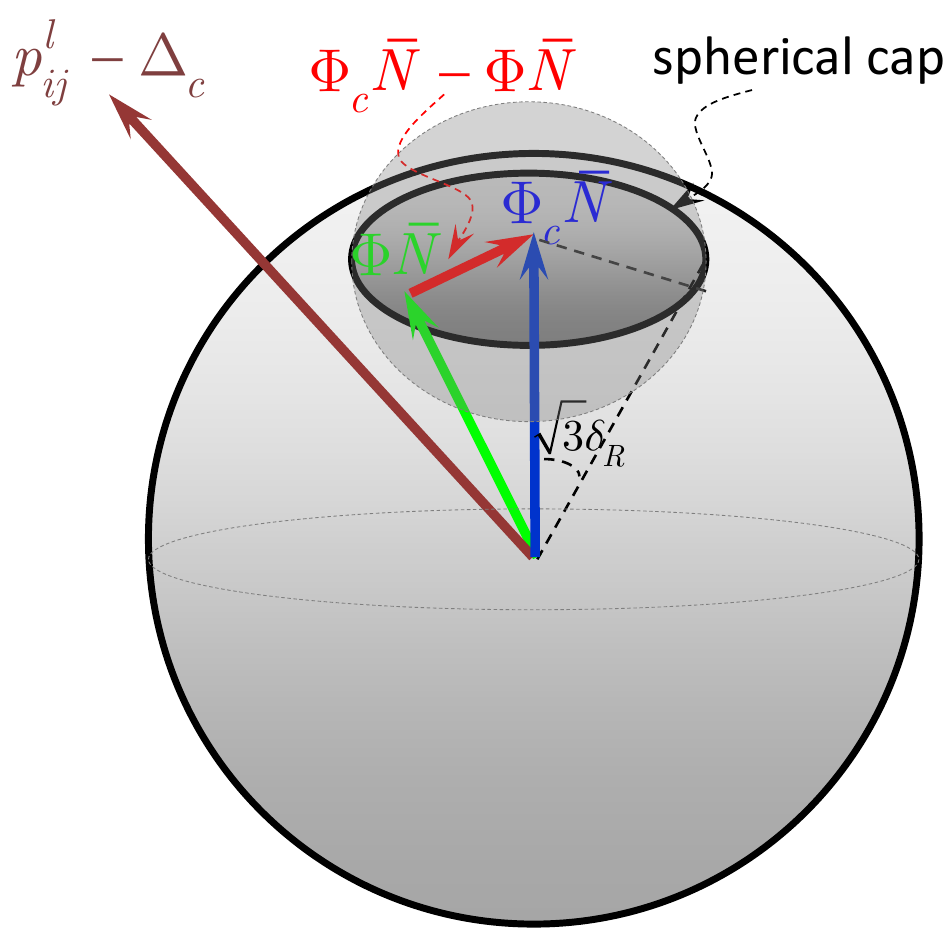}
	\vspace{-2mm}
	\caption{The vector $\Phi \bar N$ belongs to the spherical cap whose axis of symmetry is $\Phi_c \bar N$. Computation of $\delta_{ij}$ via (\ref{eq:deltaij}) involves obtaining an upper bound for the inner product $(\Phi_c \bar N -\Phi \bar N)^\top (p^l_{ij}-\Delta_c)$ by enclosing the spherical cap with the small ball depicted in the figure. The tighter bound (\ref{eq:deltaij:tight}) obtains the upper bound directly on the spherical cap.} \label{fig:sphereicalcap}
	\vspace{-5mm}
\end{figure}

Derivation of (\ref{eq:deltaij}) relies on obtaining upper bounds of the inner products $|\bar N^\top \Phi^\top (\Delta-\Delta_c)|$ and $|(\Phi_c \bar N - \Phi \bar N  )^\top (p^l_{ij}-\Delta_c)|$ over $(\Delta,\Phi) \in \mathbb{B}_t \times \mathbb{B}_R$. These upper bounds are obtained in (\ref{appendix:proof:lem:Qhat:tmp2}) by bounding the inner products with the product of the magnitude of the involved vectors, without considering the angle between these vectors. This yields a reasonably tight result for the inner product $|\bar N^\top \Phi^\top (\Delta-\Delta_c)|$ as the vector $\Delta-\Delta_c$ belongs to the 3D box $\mathbb{B}(\delta_t)$ and inevitably there exist some $\Delta$ in the box for which the angle between $\Delta-\Delta_c$ and $\Phi_c N$ is zero. For the inner product $(\Phi_c \bar N - \Phi \bar N  )^\top (p^l_{ij}-\Delta_c)$, however, the angle between the involved vectors might not be zero (or even close to zero) since $\angle (\Phi_c \bar N,\Phi \bar N) \le \sqrt{3}\delta_R$ and the vector $\Phi \bar N$ belongs to the spherical cap shown in Fig. \ref{fig:sphereicalcap} (see \cite{hartley2009global}). In fact, the upper bound (\ref{eq:deltaij}) is obtained by bounding the spherical cap with a ball with the radius $\sqrt{2(1-\cos(\sqrt{3}\delta_R))}$. Here, we obtain a tighter upper bound by maximizing the inner product directly on the spherical cap, taking into account the angle between $\Phi_c \bar N - \Phi \bar N$ and $p^l_{ij}-\Delta_c$. We have
\begin{align}
\label{eq:Phi:bound:tmp1} |(\Phi_c \bar N - \Phi \bar N  )^\top (p^l_{ij}-\Delta_c)|  = |\text{const} - (\Phi \bar N  )^\top (p^l_{ij}-\Delta_c)|.
\end{align}
where $\text{const} :=(\Phi_c \bar N)^\top (p^l_{ij}-\Delta_c)$. Hence, the particular value of $\Phi \in \mathbb{B}_R$ that maximizes (\ref{eq:Phi:bound:tmp1}) corresponds to the value that either minimizes or maximizes $g(\Phi):=(\Phi \bar N  )^\top (p^l_{ij}-\Delta_c)$. Algorithm \ref{alg:extremum:tight} provides a fast approach for computing the maximum and minimum values of $g(\Phi)$, denoted by $\gmax$ and $\gmin$, respectively (proof of this algorithm is given in Appendix \ref{appendix:Phi:bound}).  
Using Algorithm \ref{alg:extremum:tight}, a tight upper bound for (\ref{eq:Phi:bound:tmp1}) is obtained as
$|(\Phi_c \bar N - \Phi \bar N  )^\top (p^l_{ij}-\Delta_c)| \le \max(|\text{const} - \gmin|,|\text{const} - \gmax|)$.
In order to implement this new upper bound instead of the upper bound proposed in Section \ref{sec:upper:bound:function}, one only needs to replace (\ref{eq:deltaij}) with the following equation (the rest of the Algorithm \ref{alg:BnB} remains unchanged). 
\begin{align}
\label{eq:deltaij:tight} &\delta_{ij}=\sqrt{3} \delta_t+\max(|\text{const} - \gmin|,|\text{const} - \gmax|).
\end{align}
Similar to (\ref{eq:deltaij}), the new upper bound (\ref{eq:deltaij:tight}) still depends on the laser points $p^l_{ij}$ and needs to be computed for each laser point separately. It is straight-forward to show that the tight upper bound $\hat Q$ based on (\ref{eq:deltaij:tight}) also satisfies the requirements of Lemma \ref{lem:Qhat}. We provide numerical comparison of the bound (\ref{eq:deltaij}) and (\ref{eq:deltaij:tight}) in Section \ref{sec:simulation} (see Fig. \ref{fig:Qsynthetic}).
\begin{algorithm}
	\caption{Fast computation of the extremums of $g(\Phi)$}
	\textbf{Require:} the normal of the calibration plane $N$, the laser point $p^l_{ij}$, the center of the box $\mathbb{B}_t$ given by $\Delta_c$, the center and half side length of the box $\mathbb{B}_R$ given by $\Phi_c$ and $\delta_R$.
	\label{alg:extremum:tight}
	\begin{algorithmic}[1]
		\State Compute $\beta= \angle(\Phi_c \bar N,p^l_{ij}-\Delta_c)$.
		\If{$\beta \le \sqrt{3}\delta_R$}
		$\gmax=\|p^l_{ij} - \Delta_c\|$.
		\Else
		\State $\gmax\!\!=\!\!\|p^l_{ij} \!-\! \Delta_c\|\max(\cos(\beta-\!\sqrt{3}\delta_R),\cos(\beta+\sqrt{3}\delta_R)\!)$.
		\EndIf
		\If{$\beta \ge \pi-\sqrt{3}\delta_R$}
		$\gmin=-\|p^l_{ij} - \Delta_c\|$.
		\Else
		\State $\gmin\!\!=\!\!\|p^l_{ij} \!-\! \Delta_c\|\min(\cos(\beta-\!\sqrt{3}\delta_R),\cos(\beta+\sqrt{3}\delta_R)\!)$.
		\EndIf
		\State \textbf{return:} $\gmax$ and $\gmin$.
	\end{algorithmic}
\end{algorithm}
\begin{remark} \label{rem:multiple:checkerboards}
	Multiple checkerboards are sometimes used for camera-laser calibration \cite{geiger2012automatic}. In this case, detecting the inliers is harder as it is not known which one of the laser points in a given image-laser scan pair is associated with which one of the checkerboards. Assume that $k_i$ checkerboards are visible in the $i$-th image and denote the normal vectors associated with the $k$-th checkerboard ($k\in\{1,\ldots,k_i\}$) by $N^x_{ik_i}$, $N^y_{ik_i}$, and $N^z_{ik_i}$. The number of observed checkerboards in each image can be different and it is allowed that some of the checkerboards do not fall within the field of view of the camera or laser scanner in some image-laser pairs. We extend the objective function (\ref{eq:Q}) to
	\begin{align}
	\nonumber &Q(\Phi,\Delta) =\\
	\nonumber & \sum_{i=1}^{n} \!\sum_{j=1}^{m} \max_{k\in k_i}\!\Big(\! \lfloor |{\bar N{\hspace{-0.3mm}{}^{x}_{ik_i}}}^{\hspace{-2.5mm}\top} \Phi^\top (p^l_{ij}-\Delta)-\| N{\hspace{-0.3mm}{}^{x}_{ik_i}}\| | < d_x + \epsilon \rfloor\\
	\nonumber &~~~~~~~~~~~~~~~~ \times\! \lfloor |{\bar N{\hspace{-0.3mm}{}^{y}_{ik_i}}}^{\hspace{-2.5mm}\top} \!\Phi^{\!\top} (p^l_{ij}-\Delta)-\| N{\hspace{-0.3mm}{}^{y}_{ik_i}}\|| < d_y + \epsilon \rfloor \\
	\label{eq:Q:multicheckerboard} &~~~~~~~~~~~~~~~~ \times\! \lfloor |{\bar N{\hspace{-0.3mm}{}^{z}_{ik_i}}}^{\hspace{-2.5mm}\top} \!\Phi{\!^\top} (p^l_{ij}-\Delta)-\| N{\hspace{-0.3mm}{}^{z}_{ik_i}}\|| < \epsilon \rfloor \Big).
	\end{align}
	Using the $\max$ operator where the correspondence between the measurements are unknown has been successfully practiced in the literature (see e.g. \cite{breuel2003implementation,parra2014fast}). For a given calibration $(\Phi,\Delta)$, the objective function (\ref{eq:Q:multicheckerboard}) choses the correspondence between the checkerboards and the laser scans that yields the maximum number of inliers. Note that extending the objective function to (\ref{eq:Q:multicheckerboard}) does not change the structure of Algorithm \ref{alg:BnB} at all and the results of Lemma \ref{lem:Qhat} still hold. One only needs to compute $Q(\Phi_c,\Delta_c)$ and $\hat Q(\mathbb{B}_R,\mathbb{B}_t)$ using (\ref{eq:Q:multicheckerboard}) instead of (\ref{eq:Q}). The optimum correspondence between the laser points and the checkerboards in each image is obtained at line \ref{alg:line:inliers} of Algorithm \ref{alg:BnB}.\carrew
\end{remark}

\begin{figure*}
	\centering
	\begin{subfigure}[b]{.33\textwidth}
		\includegraphics[width=\columnwidth]{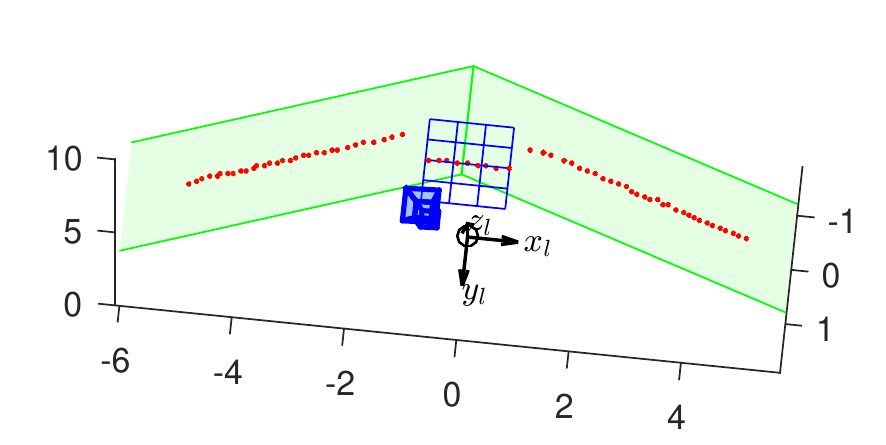}
		\vspace{-8mm}
		\caption{}
		\label{fig:synthetic:3da}
	\end{subfigure}
	\begin{subfigure}[b]{.32\textwidth}
		\includegraphics[width=\columnwidth]{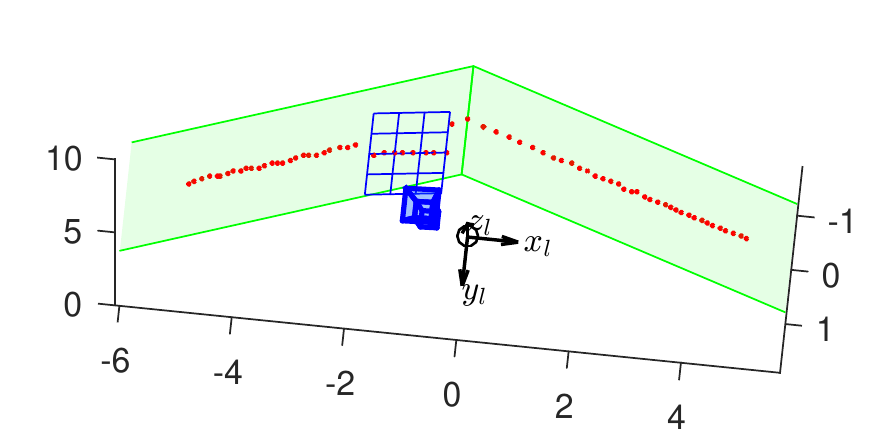}
		\vspace{-8mm}
		\caption{}
		\label{fig:synthetic:3db}
	\end{subfigure}
	\begin{subfigure}[b]{.33\textwidth}
		\includegraphics[width=\columnwidth]{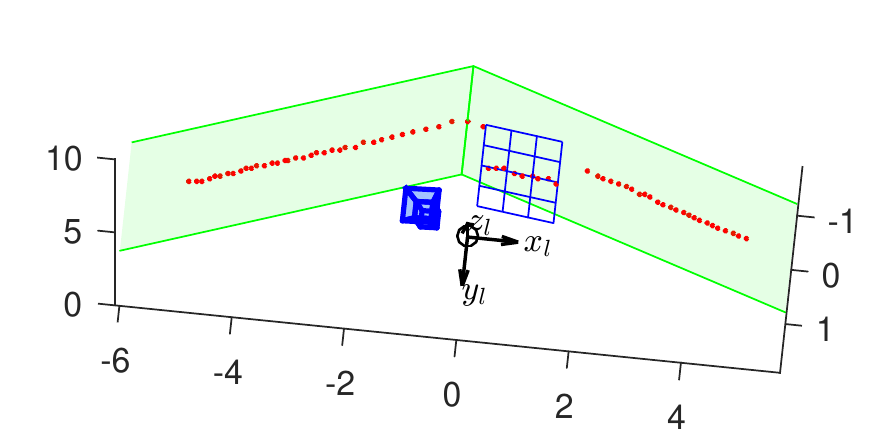}
		\vspace{-8mm}
		\caption{}
		\label{fig:synthetic:3dc}
	\end{subfigure}
	\\
	\begin{subfigure}[b]{.33\textwidth}
		\includegraphics[width=\columnwidth]{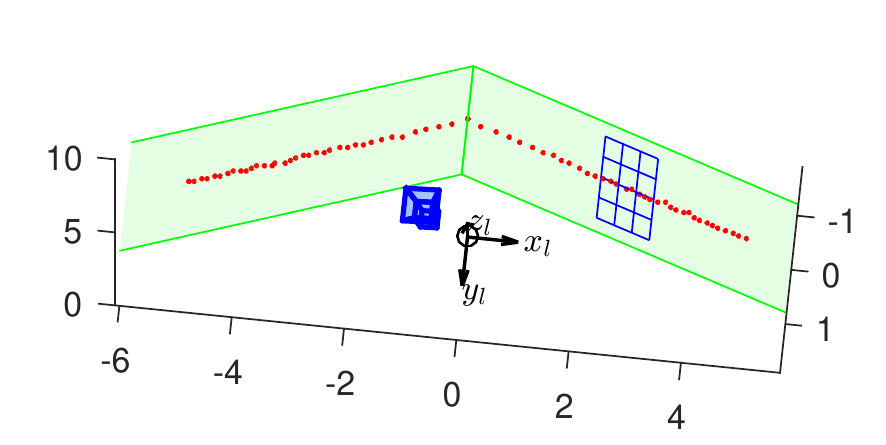}
		\vspace{-8mm}
		\caption{}
		\label{fig:synthetic:3de}
	\end{subfigure}
	\begin{subfigure}[b]{.32\textwidth}
		\includegraphics[width=\columnwidth]{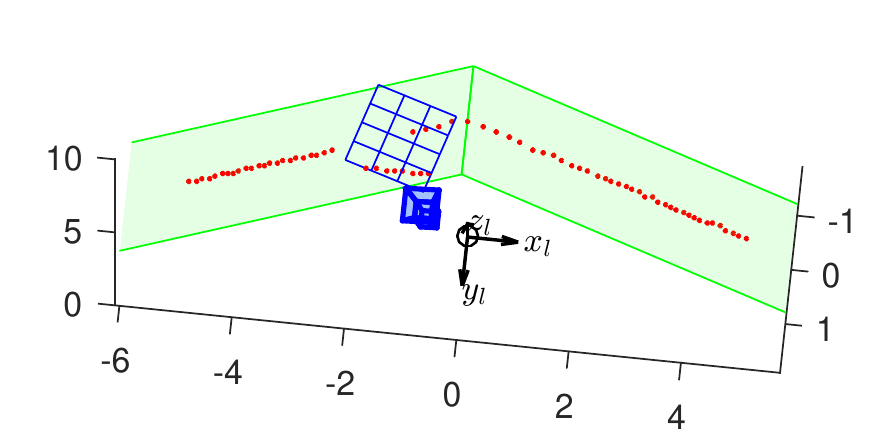}
		\vspace{-8mm}
		\caption{}
		\label{fig:synthetic:3df}
	\end{subfigure}
	\begin{subfigure}[b]{.33\textwidth}
		\includegraphics[width=\columnwidth]{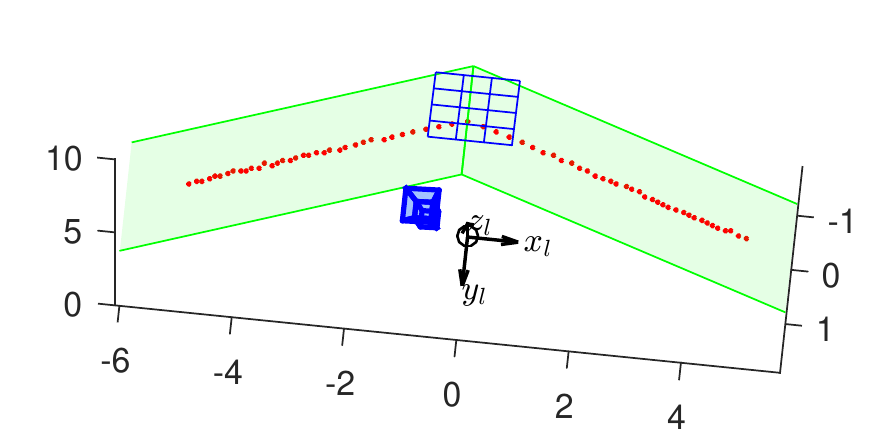}
		\vspace{-8mm}
		\caption{}
		\label{fig:synthetic:3dg}
	\end{subfigure}
	\\
	\begin{subfigure}[b]{.33\textwidth}
		\includegraphics[width=\columnwidth]{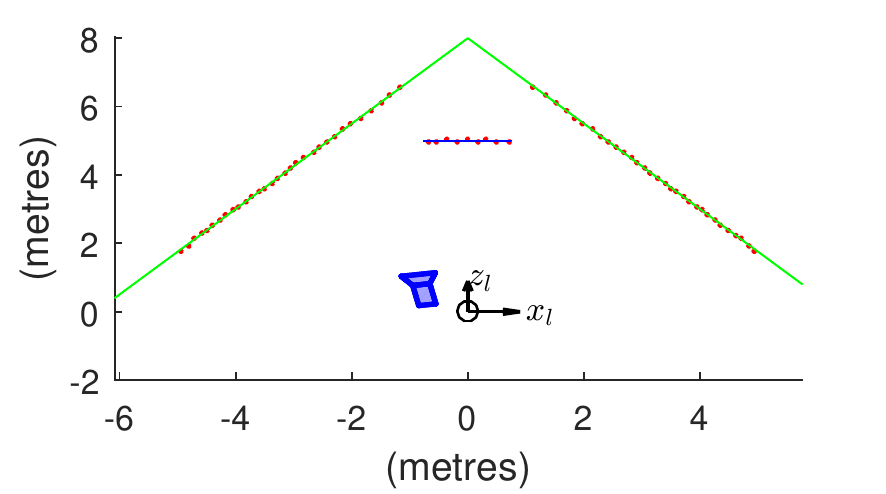}
		\vspace{-7mm}
		\caption{}
		\label{fig:synthetic:3dh}
	\end{subfigure}
	\begin{subfigure}[b]{.32\textwidth}
		\includegraphics[width=\columnwidth]{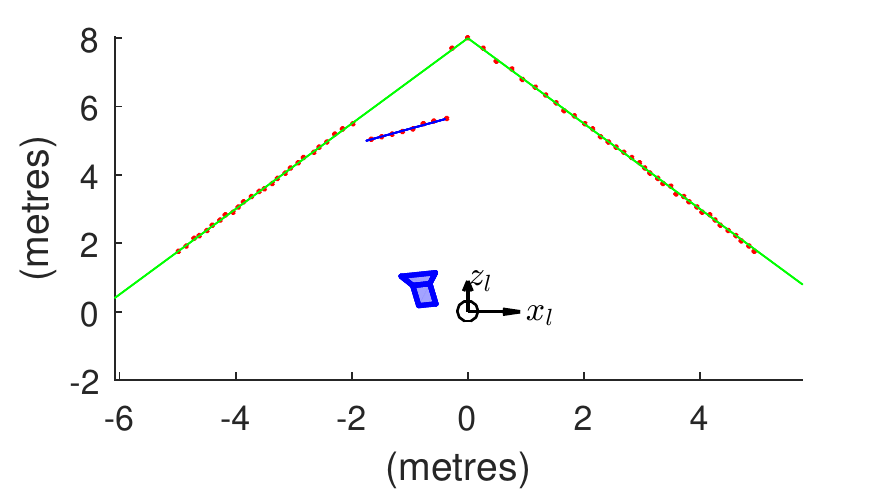}
		\vspace{-7mm}
		\caption{}
		\label{fig:synthetic:3di}
	\end{subfigure}
	\begin{subfigure}[b]{.33\textwidth}
		\includegraphics[width=\columnwidth]{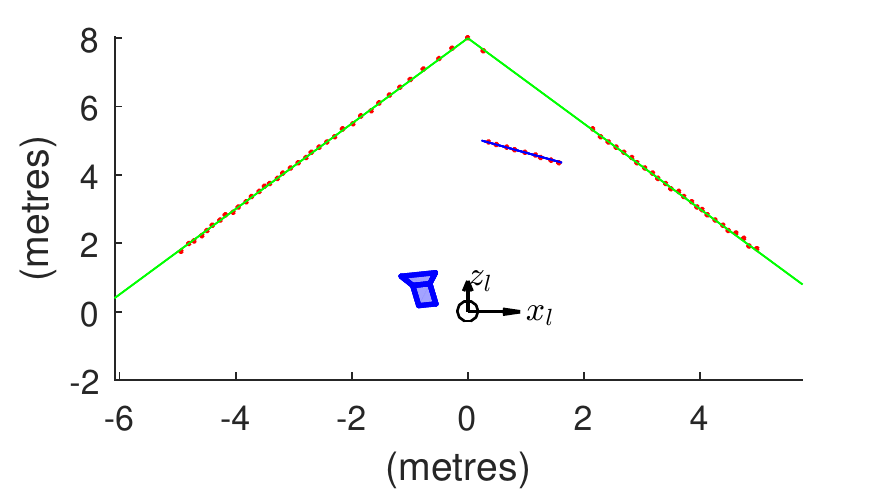}
		\vspace{-7mm}
		\caption{}
		\label{fig:synthetic:3dj}
	\end{subfigure}
	\\
	\begin{subfigure}[b]{.33\textwidth}
		\includegraphics[width=\columnwidth]{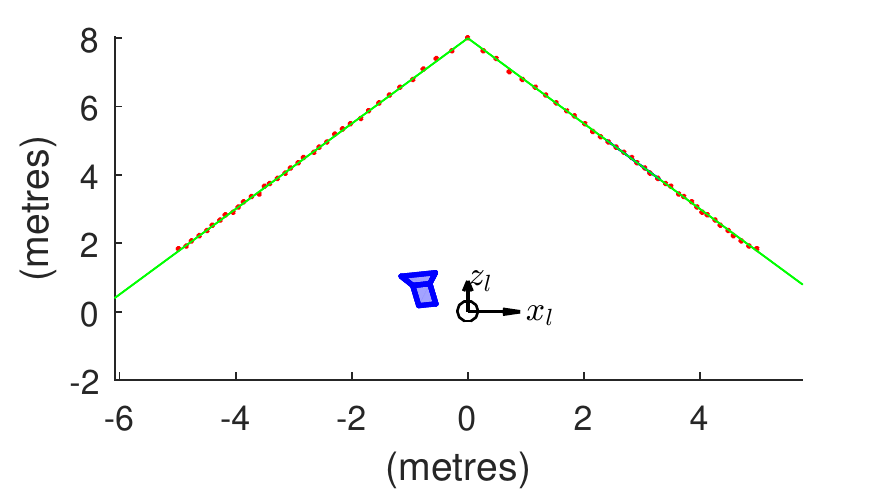}
		\vspace{-7mm}
		\caption{}
		\label{fig:synthetic:3dk}
	\end{subfigure}
	\begin{subfigure}[b]{.32\textwidth}
		\includegraphics[width=\columnwidth]{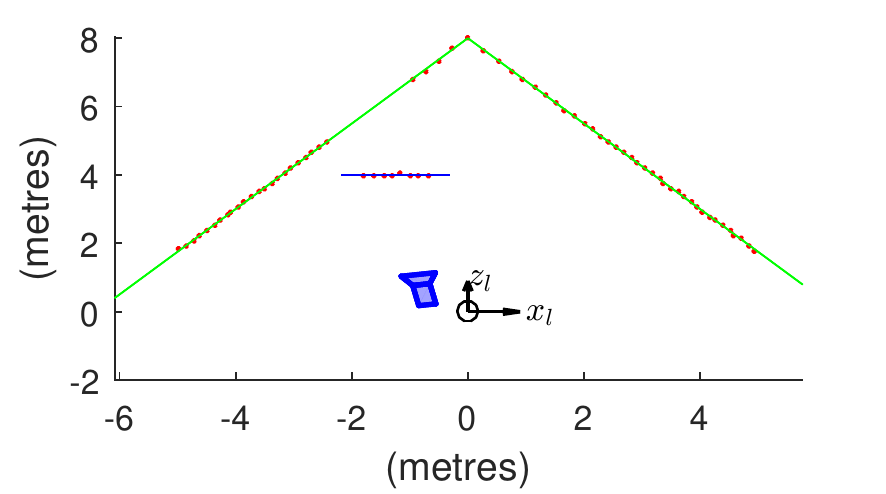}
		\vspace{-7mm}
		\caption{}
		\label{fig:synthetic:3dl}
	\end{subfigure}
	\begin{subfigure}[b]{.33\textwidth}
		\includegraphics[width=\columnwidth]{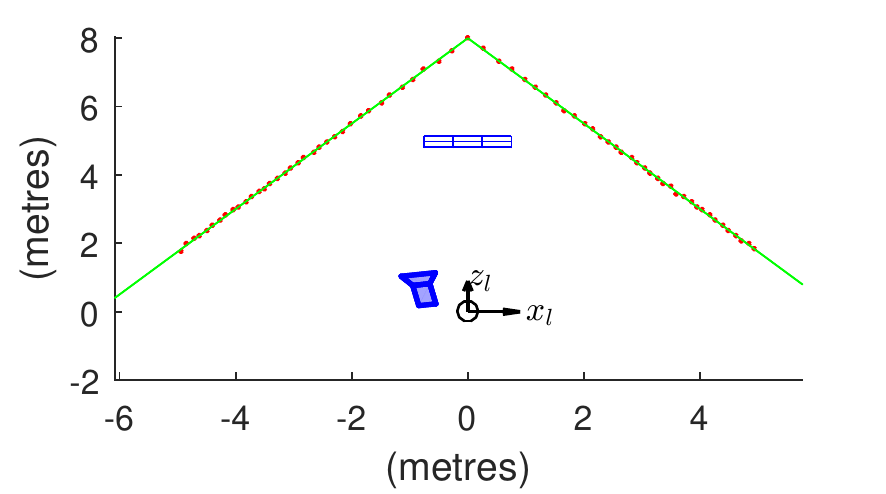}
		\vspace{-7mm}
		\caption{}
		\label{fig:synthetic:3dm}
	\end{subfigure}	
	\vspace{-2mm}
	\caption{Synthetic data used for the simulation example of Section \ref{sec:simulation}. The top six figures are 3D plots and the bottom six figures show their corresponding top views. The laser points are depicted by red dots, expressed in the laser coordinate frame depicted by a black coordinate frame whose axes are labeled as $x_l$, $y_l$, $z_l$. The camera is depicted by a blue camera sign, the checkerboard is depicted by a blue mesh surface, and the walls are shown in green. The checkerboard in Fig. \ref{fig:synthetic:3de} is placed on a wall preventing any discontinuity of the range measurements. The laser scans only hit a part of the checkerboard in Fig. \ref{fig:synthetic:3df} and they do not hit the checkerboard at all in Fig. \ref{fig:synthetic:3dg}.}
	\label{fig:synthetic:3d2d}
	\vspace{-5mm}
\end{figure*}
\section{Simulation results} \label{sec:simulation}
\begin{figure} 
	\includegraphics[width=9cm]{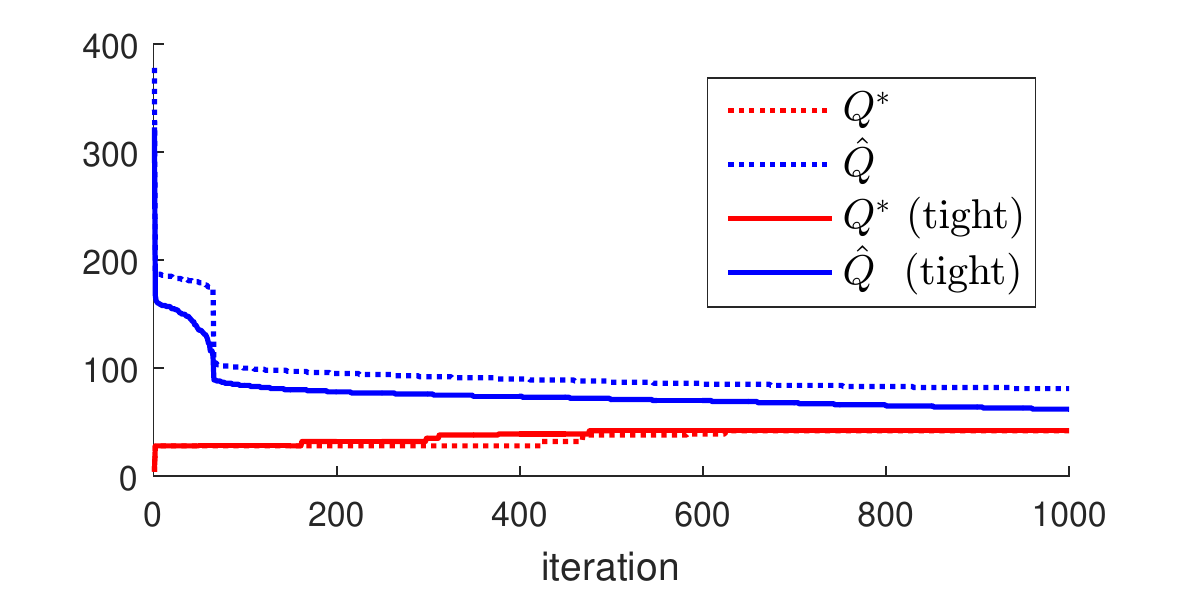}
	\vspace{-2mm}
	\caption{Evolution of $Q^\star$ and the upper bound $\hat Q$ for the simulations of Section \ref{sec:simulation}. Dotted lines and solid lines, respectively, correspond to the case where the upper bound based on (\ref{eq:deltaij}) and (\ref{eq:deltaij:tight}) are employed.} \label{fig:Qsynthetic}
	\vspace{-5mm}
\end{figure}
\begin{figure} 
	\centering
	\includegraphics[width=7cm]{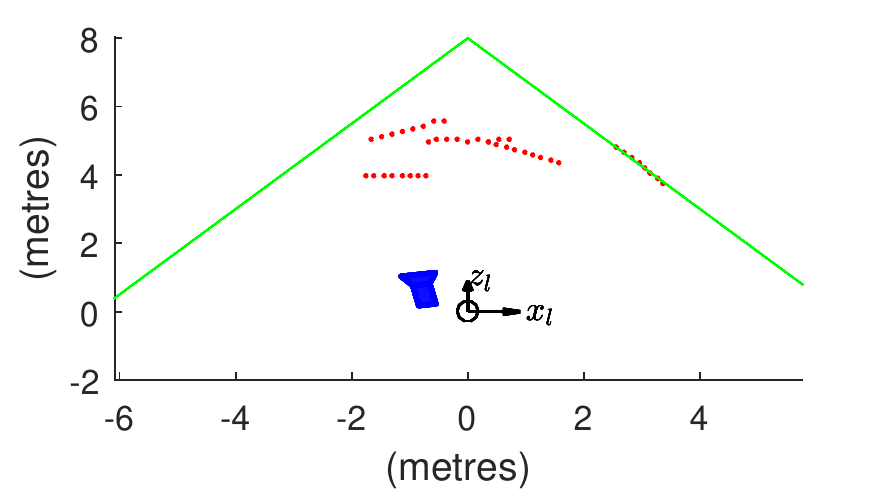}
		\vspace{-2mm}
	\caption{Red dots are the inliers laser points identified by the BnB method (cf. Fig. \ref{fig:synthetic:3dh} to Fig. \ref{fig:synthetic:3dm}).} \label{fig:inliersBnBsynthetic}
		\vspace{-5mm}
\end{figure}
This section provides simulation studies demonstrating that the proposed BnB method is able to effectively extract the checkerboards even in challenging situations where some of the checkerboards are placed on the walls (hence do not create discontinuity in range measurements), or where some of the checkerboards either partially hit or are not hit at all by laser scans in some of the image-laser scan pairs.

We choose the extrinsic rotation $\Phi$ to be a rotation of $10$ degrees around the $y$-axis of the laser scanner and we choose the extrinsic translation $\Delta=[-0.75~-0.2~0.5]^\top$. A $1.5 \text{(m)} \times 1.5 \text{(m)}$ checkerboard is placed in $6$ different positions in front of the camera and laser scanner, each time with a different orientation (see Fig. \ref{fig:synthetic:3d2d}). We consider a 2D laser scanner providing range measurements in a horizontal plane, every $2$ degrees from $-70$ degrees to $70$ degrees yielding a total of $71$ measured laser points in each image. The laser-camera setup is placed in a triangular shaped room where the distance of the walls to the laser scanner is $5$ meters and the intersection line of the walls is $8$ meters away from the laser scanner. The laser points that do not hit the checkerboards will hit the walls. A uniformly distributed random noise between $-2$ centimeters and $2$ centimeters is added to the resulting range measurements.

The $6$ synthetically generated sets are depicted in Fig. \ref{fig:synthetic:3d2d}. To make the data more challenging for checkerboard extraction, in Fig. \ref{fig:synthetic:3de}, the checkerboard is placed exactly on the wall, preventing any discontinuity in the range measurements. Also, we choose the checkerboard orientation such that the laser scans only partially hit it in Fig. \ref{fig:synthetic:3df} and do not hit it at all in Fig. \ref{fig:synthetic:3dg}. The total of $6*71=426$ laser points are measured, out of which $42$ points correspond to the checkerboards (inliers) and the rest fall on the walls. Checkerboard normals represented in the camera frame are polluted by a uniformly distributed random rotation of maximum $1$ degrees along each axis of the camera to model the camera noise. The resulting checkerboard normals along with the laser scan points are passed to Algorithm \ref{alg:BnB}. The threshold $\epsilon$ is chosen to be $7$ centimeters and the initial search space is chosen as $\mathbb{B}_R=\mathbb{B}(\frac{\pi}{12})$ and $\mathbb{B}_t=\mathbb{B}(1~\text{meter})$.

Fig. \ref{fig:Qsynthetic} shows the evolution of $Q^\star$ and $\hat Q$ for the first $1000$ iterations of BnB when they are examined at line \ref{alg:line:Qstar} of Algorithm \ref{alg:BnB}. Fig. \ref{fig:Qsynthetic} includes both the case where the original bound (\ref{eq:deltaij}) or the tighter bound (\ref{eq:deltaij:tight}) is used in BnB. The upper bounds initially decay very fast, but their rate of decay decreases for large iterations (this is a typical characteristic of objective functions designed using the floor operator). As expected, $\hat Q$ computed using (\ref{eq:deltaij:tight}) decays faster than that of (\ref{eq:deltaij}), showing that (\ref{eq:deltaij:tight}) indeed yields a tighter upper bound. The BnB method is able to find all of the inliers after $475$ iterations when the tight upper bound is used and after $625$ iterations when the original upper bound is used. The runtime is reported in Section \ref{sec:experiment}. Observe that $\hat Q$ still decays at the $1000$-th iteration and will eventually reach $Q^\star$ and the algorithm terminates. It is a common practice to terminate BnB algorithms after large enough number of inliers are detected and $Q^\star$ stops growing for a large enough consecutive iterations. Fig. \ref{fig:inliersBnBsynthetic} shows the inliers detected at line \ref{alg:line:inliers} of Algorithm \ref{alg:BnB} (for both the bound (\ref{eq:deltaij}) and (\ref{eq:deltaij:tight}) demonstrating that all of the $42$ inliers are successfully detected.

\section{Experimental results} \label{sec:experiment}
\begin{figure} 
	\centering
	\includegraphics[width=6cm]{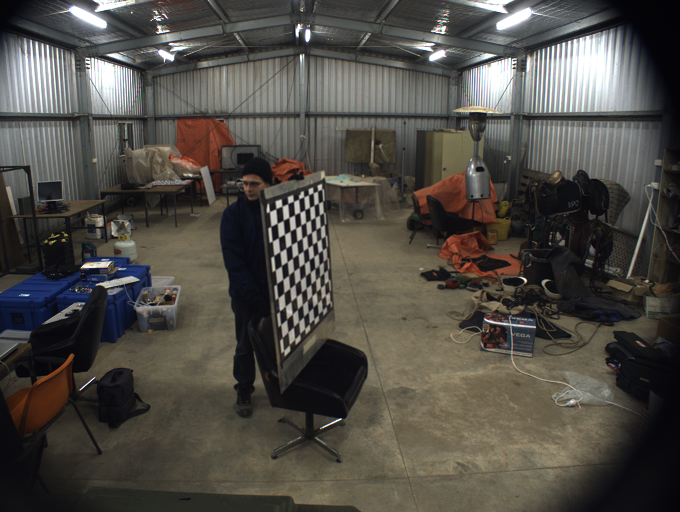}
	\vspace{-5mm}
	
	\includegraphics[width=8cm]{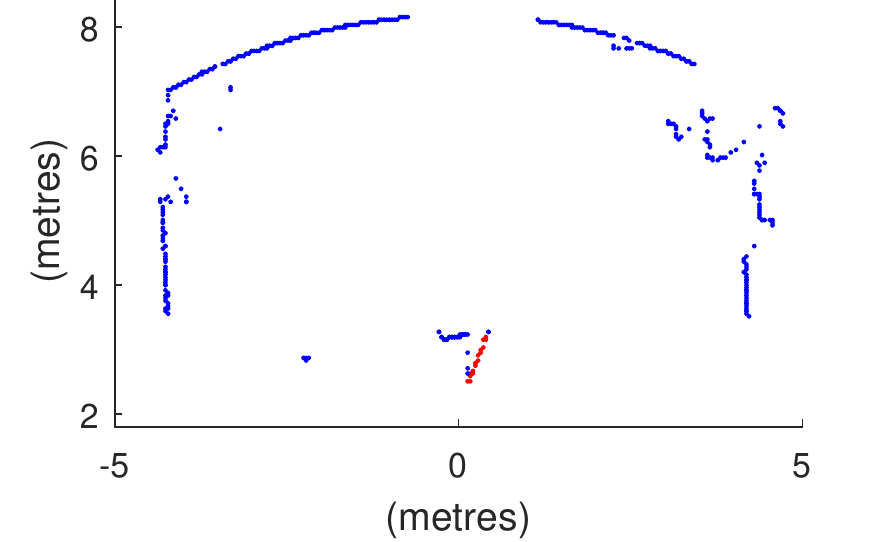}
	\vspace{-1mm}
	\caption{A sample image (top) and its corresponding laser measurements (bottom). The inliers and outliers are identified by our BnB method are colored red and blue, respectively. Image and laser data from \cite{kassir2010reliable}.} \label{fig:inliersBnBexperimentSample}
	\vspace{-3mm}
\end{figure}

In this Section, we verify the effectiveness of the proposed BnB method using real images and 2D laser scan dataset of \cite{kassir2010reliable}\footnote{Available online: \url{http://www-personal.acfr.usyd.edu.au/akas9185/AutoCalib/}}. The dataset contains $20$ images together with their corresponding laser scans. Each laser scan contains $401$ range measurements in the horizontal plane, measured every $0.25$ degrees from $-50$ degrees to $50$ degrees. A checkerboard is placed in a different place in each image. The images are processed using the camera calibration toolbox \cite{bouguet2010camera}. This toolbox extracts out the corners of the checkerboard, computes camera intrinsics and lens distortion parameters, and provides extrinsic transformation of each checkerboard plane wrt. to the camera frame. The output of \cite{bouguet2010camera} is used both in our BnB method and in the toolbox of \cite{kassir2010reliable}. Having the output of the camera calibration toolbox, it is straight-forward to compute the checkerboard normals $N^x_i$, $N^y_i$, and $N^z_i$ required by Algorithm \ref{alg:BnB}. The dimensions of the checkerboard rectangles are provided in the dataset, but the dimensions of the calibration board itself (which are larger than the area of the checkerboard, see Fig. \ref{fig:inliersBnBexperimentSample}) are not provided. From the images, we estimate the dimensions of the calibration board to be at least $0.83 (m) \times 0.83 (m)$ and we use these values as $2 d_x$ and $2 d_y$ in Algorithm \ref{alg:BnB}. The search space is initialized to $\mathbb{B}_R=\mathbb{B}(\frac{\pi}{18})$ and $\mathbb{B}_t=\mathbb{B}(0.5)$ and the inlier thresholds is chosen as $\epsilon=0.1$ meters\footnote{The search space is initialized large enough to contain the true calibration parameters $\Phi$ and $\Delta$. By estimating the calibration parameters via the toolbox of \cite{kassir2010reliable}, we verified that the angle of rotation corresponding to $\Phi$ is less than $5$ degrees and the amplitude of the transformation is less than $0.5$ meters along each axis. Algorithm \ref{alg:BnB} works if a larger search space is chosen, although this increases the number of iterations.}.

\begin{figure} 
	\includegraphics[width=8cm]{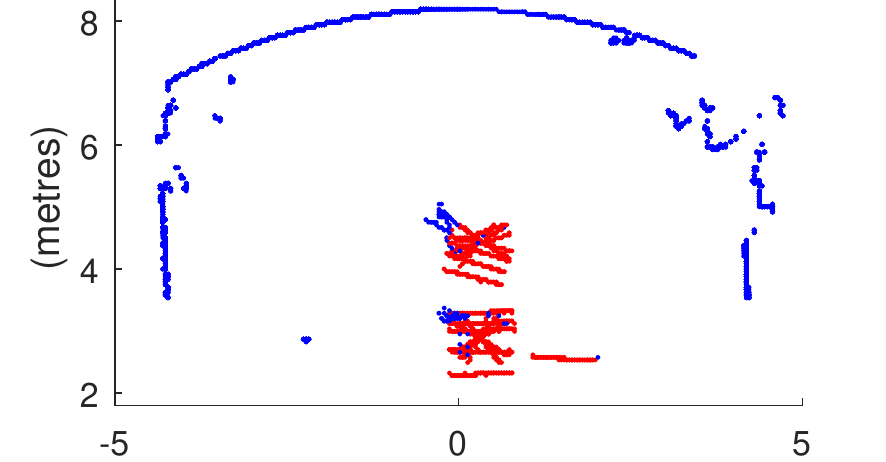}		
	\includegraphics[width=8cm]{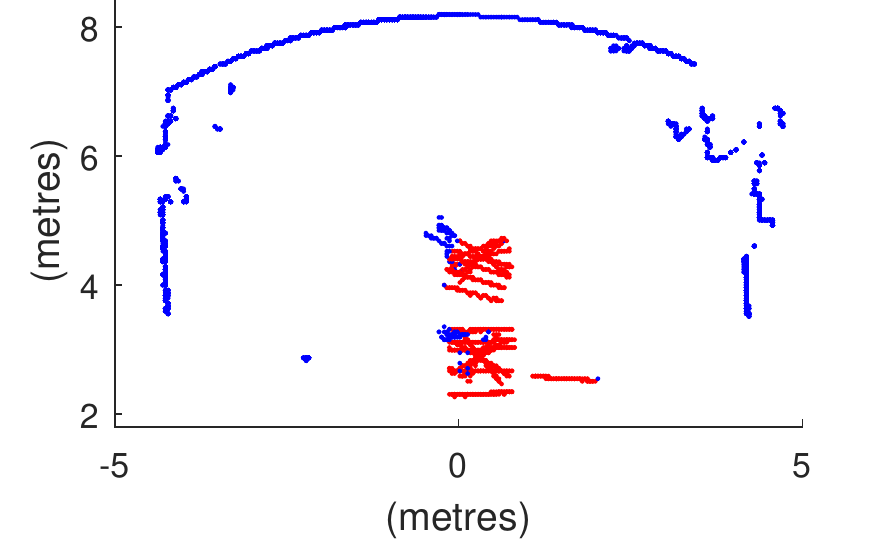}
	\vspace{-2mm}
	\caption{The laser points identified as inliers by the BnB method (top plot) and by the toolbox of \cite{kassir2010reliable} (bottom plot) are colored red. The rest of the laser points (outliers) are colored blue. The laser scanner is placed at the origin.} \label{fig:inliersBnBexperiment}
	\vspace{-2mm}
\end{figure}

Fig. \ref{fig:inliersBnBexperiment} shows the extracted checkerboards after the first $1000$ iterations of our BnB method with the tight bound (top plot) versus the checkerboards extracted by the toolbox of \cite{kassir2010reliable}\footnote{We adjusted the line line extraction thresholds of\cite{kassir2010reliable} to detect all of the checkerboards. Otherwise, one of the checkerboards would not be detected.}. The total of $1038$ inlier laser points are extracted by our method and $1036$ inliers are detected by the toolbox of \cite{kassir2010reliable}. The extracted inlier laser points of both methods are almost identical. No ground truth inliers are given in the data set. Nevertheless, since the employed data has been verified to work efficiently with the toolbox of \cite{kassir2010reliable}, the validity of the extracted inliers of the BnB method is verified. Fig. \ref{fig:inliersBnBexperimentSample} shows a sample of the image-laser scan pairs.  Some of the points near the edges of the checkerboard fall on the hand or clothes of the person holding the checkerboard, and are correctly classified as outliers by Algorithm \ref{alg:BnB}. Also, since $d_x$ and $d_y$ are chosen smaller than the true dimensions of the checkerboard, occasionally one or two inlier laser points at the edge of the checkerboard are classified as outliers by our method. Same effect is observed in the results of \cite{kassir2010reliable}. This is not important in practice as those laser points are naturally more noisy and might not be very helpful in calibration \cite{boehler2003investigating}.

On our machine equipped with a Core i7-4720HQ processor, the total $1000$ iterations of the BnB takes $249$ seconds while the toolbox of \cite{kassir2010reliable} extracts the checkerboards in less than $2$ seconds.
We emphasis, however, that the calibration is a one-off process and the run time of BnB is negligible compared to the data acquisition, justifying its generality, flexibility with practical conditions, and robustness gains. We emphasis that \cite{kassir2010reliable} is tailored to 2D laser scanners, necessarily requires stationary background, and relies on range discontinuities, all of which are satisfied in the employed dataset. Our BnB method does not impose any of these assumption and is applicable to 3D laser scanners as well.


\section{Conclusion} \label{sec:conclusion}
We formulate the checkerboard extraction as a combinatorial optimization problem with a clear cut objective function and we propose a branch-and-bound technique for optimizing the objective. The proposed BnB method is able to robustly extract the checkerboard in a diverse range of practical scenarios, including but not limited to where either 2D or 3D laser scanners are used, multiple checkerboards are present in the scene, checkerboards are placed on the walls and no range discontinuity is associated with checkerboard edges, some of the checkerboards are only partially hit or are not hit at all by the laser scans, background changes from one scan to another, and multiple unwanted objects are present in the scene creating undesired range discontinuities. We demonstrate effective application of the proposed method via simulation and experimental studies.


\appendices 
\vspace{-2mm}
\section{} \label{appendix:Phi:bound}
\begin{figure} 
	\centering
	\includegraphics[width=7cm]{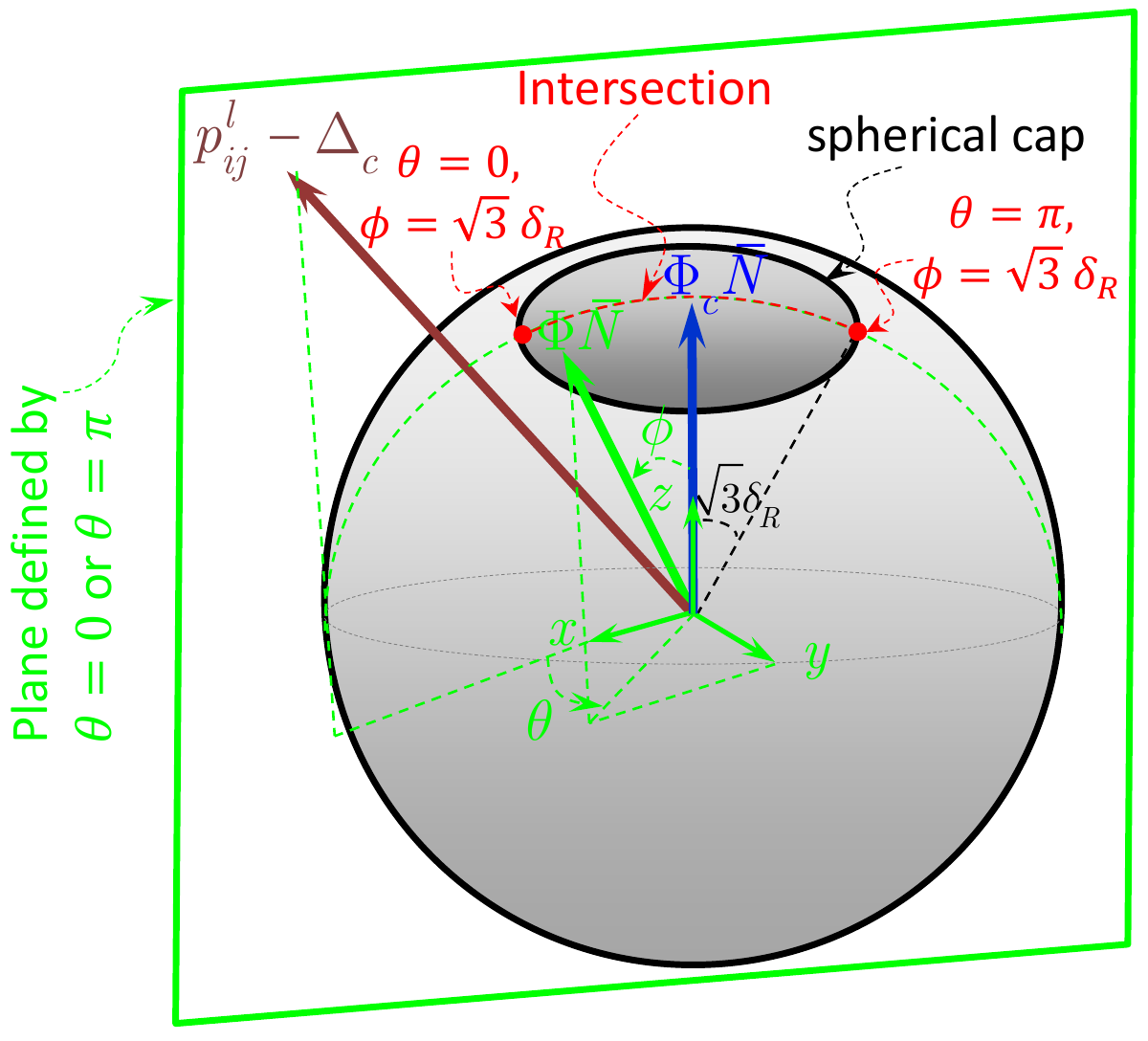}
	\vspace{-2mm}
	\caption{The dashed red curve shows the intersection of the plane $\theta\!=\!0$ (or $\theta\!=\!\pi$) with the spherical cap. The intersection of the plane with the boundary of the spherical cap corresponds to the points $(\theta\!=\!0,\phi\!=\!\sqrt{3}\delta_R)$ and $(\theta\!=\!\pi,\phi\!=\!\sqrt{3}\delta_R)$.} \label{fig:sphericalcap:intersect}
	\vspace{-6mm}
\end{figure}
Consider a coordinate frame whose center is at the origin of $\Phi_c \bar N$ and its $z$-axis is along $\Phi_c \bar N$ (see Fig. \ref{fig:sphericalcap:intersect}). Choose the $x$-axis of this coordinate frame such that it falls within the plane containing $\bar N$ and $p^l_{ij}-\Delta_c$. The $y$-axis is defined by $z \times x$. Any vector $\Phi \bar N \in \mathbb{B}_R$ is expressed in the $xyz$ coordinate as $\Phi \bar N \!=\! [\sin\phi\cos \theta,\sin\phi\sin\theta,\cos\phi]^\top$ where $\phi$ is the angle between the $z$-axis $\Phi \bar N$, and $\theta$ is the angle between the $x$-axis and the projection of $\Phi \bar N$ into the $xy$ plane. Similarly, the expression of $ p^l_{ij}-\Delta_c$ is given by $p^l_{ij}-\Delta_c=\| p^l_{ij}-\Delta_c \|[\sin\beta,0,\cos\beta]^\top$ where $\beta\!:=\!\angle(\Phi_c \bar N,p^l_{ij}-\Delta_c)$. We have
\begin{align}
\nonumber g(\Phi)&=g(\theta,\phi):=(\Phi \bar N  )^\top (p^l_{ij}-\Delta_c) \\
\label{eq:Phi:bound:tmp2} &= \| p^l_{ij}-\Delta_c \| (\sin\beta \sin\phi \cos\theta + \cos\beta\cos\phi).
\end{align}
Finding the value of $\Phi$ that maximizes or minimizes (\ref{eq:Phi:bound:tmp2}) boils down to maximizing/minimizing (\ref{eq:Phi:bound:tmp2}) wrt. $\theta$ and $\phi$. Observe that $0 \le \theta \le 2\pi$ and $0 \le \phi \le \sqrt{3}\delta_R$. We have $\frac{\partial g(\theta,\phi)}{\partial \theta} = 0 \Rightarrow  \sin\theta=0 \Rightarrow \theta=0~\text{or}~\pi$. This implies that extremums of (\ref{eq:Phi:bound:tmp2}) occur on the intersection of the plane that includes $\Phi_c \bar N$ and $p^l_{ij}-\Delta_c$ (this plane is characterized by $\theta=0$ or $\theta=\pi$) and the spherical cap of Fig. \ref{fig:sphereicalcap} (the intersection is depicted in Fig. \ref{fig:sphericalcap:intersect}). Replacing for $\theta=0$ or $\theta=\pi$ in (\ref{eq:Phi:bound:tmp2}), we have $g(0,\phi) = \| p^l_{ij}-\Delta_c \| \cos(\beta-\phi)$ and $g(\pi,\phi) = \| p^l_{ij}-\Delta_c \| \cos(\beta+\phi)$. Hence,
$\frac{\partial g(0,\phi)}{\partial \phi} = \| p^l_{ij}-\Delta_c \| \sin(\beta-\phi)$ and $\frac{\partial g(\pi,\phi)}{\partial \phi} = -\| p^l_{ij}-\Delta_c \| \sin(\beta+\phi)$. 
Observe that $\frac{\partial g(0,\phi)}{\partial \phi} = 0$ yields either $\phi=\beta$ or $\phi=\pi+\beta$. Since $0 \le \phi \le \pi$, only $\phi=\pi+\beta$ is valid. Similarly, $\frac{\partial g(\pi,\phi)}{\partial \phi} = 0$ yields either $\phi=\pi-\beta$ or $\phi=-\beta$ where $\phi=-\beta$ is invalid. If either of $\phi=\beta$ or $\phi=\pi-\beta$ creates a rotation $\Phi$ that falls within the box $\mathbb{B}_R$, this rotation corresponds to an extremum of $g$. These correspond to the cases where either $\Phi \bar N=\frac{p^l_{ij} - \Delta_c}{\|p^l_{ij} - \Delta_c\|}$, yielding the maximum of (\ref{eq:Phi:bound:tmp2}) to be $\gmax=\|p^l_{ij} - \Delta_c\|$, or $\Phi \bar N=-\frac{p^l_{ij} - \Delta_c}{\|p^l_{ij} - \Delta_c\|}$, yielding the minimum of (\ref{eq:Phi:bound:tmp2}) to be $\gmin=-\|p^l_{ij} - \Delta_c\|$. Otherwise, since $g(\theta,\phi)$ is continuous on $\phi\in[0,\sqrt{3}\delta_R]$, the extremums might happen at the boundary point $\phi=\sqrt{3}\delta_R$
, which yield the candidate extremums $g(0,\sqrt{3}\delta_R)=\|p^l_{ij} - \Delta_c\| \cos(\beta-\sqrt{3}\delta_R)$ and $g(\pi,\sqrt{3}\delta_R)=\|p^l_{ij} - \Delta_c\| \cos(\beta+\sqrt{3}\delta_R)$. Thus, in order to find the extremums, one should first check if either of $\phi=\beta$ or $\phi=\pi-\beta$ falls inside the box $\mathbb{B}_R$, yielding the maximum or minimum, respectively. Otherwise, the maximum/minimum is obtained from the boundary points. Algorithm \ref{alg:extremum:tight} summarizes this method.

\begin{algorithm}
	\caption{Branch-and-bound method for maximizing (\ref{eq:Q})}
	\textbf{Require:} $\epsilon$, $d_x$, $d_y$, $\mathbb{B}(\delta_R)$, $\mathbb{B}(\delta_t)$, $N^x_i,N^y_i,N^z_i$, $p^l_{ij}$ for $i=1,\ldots,n,j=1,\ldots, m$.
	\label{alg:BnB}
	\begin{algorithmic}[1]
		\State Initialize priority queue $q$, $\mathbb{B}_R \leftarrow \mathbb{B}(\delta_R)$, $\mathbb{B}_t \leftarrow \mathbb{B}(\delta_t)$. 
		\State Insert $(\mathbb{B}_R,\mathbb{B}_t)$ into $q$,  $\Phi_c \leftarrow$ center of $\mathbb{B}_R$, $\Delta_c \leftarrow$ center of $\mathbb{B}_t$. $Q^\star \leftarrow Q(\Phi_c,\Delta_c)$.
		\While{$q$ is not empty}
		\State Find the element of $q$ with the highest $\hat Q$ and remove this element from $q$ and insert it into $(\mathbb{B}_R,\mathbb{B}_t)$.
		\If{$\hat Q(\mathbb{B}_R,\mathbb{B}_t)=Q^\star$} terminate. \label{alg:line:Qstar} \EndIf
		\State \label{alg:line:branchR} $\mathbb{B}_{R_{m}},~m=1,\ldots,8 \leftarrow \text{branch}(\mathbb{B}_R)$.
		\State \label{alg:line:brancht} $\mathbb{B}_{t_{n}},~n=1,\ldots,8 \leftarrow \text{branch}(\mathbb{B}_t)$.
		\For{$k$=1 to $8$} 
		\For{$\ell$=1 to $8$}
		\State
		\vspace{-7mm}
		\begin{align*}
		&~~~~~~~~~~~{\Phi_c}_k \!\!\leftarrow \text{center of}~ \mathbb{B}_{R_{k}},~{\Delta_c}_\ell \leftarrow ~\text{center of}~ \mathbb{B}_{t_{\ell}}
		\end{align*}
		\vspace{-6mm}
		
		~~~~~~Compute $\hat Q(\mathbb{B}_{R_{k}},\mathbb{B}_{t_{\ell}})$ and $Q({\Phi_c}_k,{\Delta_c}_\ell)$. 
		\If{$\hat Q(\mathbb{B}_{R_{k}},\mathbb{B}_{t_{\ell}}) > Q^\star$}
		\State Insert $(\mathbb{B}_{R_{k}},\mathbb{B}_{t_{\ell}})$ into $q$.
		\If{$Q({\Phi_c}_k,{\Delta_c}_\ell) > Q^\star$}
		\State \label{alg:line:star:update}
		\vspace{-7mm}							
		\begin{align*}
		&~~~~~~Q^\star \leftarrow Q({\Phi_c}_k,{\Delta_c}_\ell),\\
		&~~~~~~(\Phi^\star,\Delta^\star) \leftarrow ({\Phi_c}_k,{\Delta_c}_\ell).
		\end{align*}
		\vspace{-6mm}
		\EndIf
		\EndIf
		\EndFor
		\EndFor
		\EndWhile
		\State $inliers^\star \leftarrow$ index of the laser points that contribute $1$ to the summation (\ref{eq:Q}) in evaluating $Q(\Phi^*,\Delta^*)$. \label{alg:line:inliers}
		\State \textbf{return} $\Phi^\star,\Delta^\star,~inliers^\star$.
	\end{algorithmic}
\end{algorithm}

\bibliographystyle{IEEEtran}
\bibliography{librarysample}

\end{document}